\documentclass[pdflatex,sn-mathphys-num]{sn-jnl}

\usepackage{graphicx}%
\usepackage{multirow}%
\usepackage{amsmath,amssymb,amsfonts}%
\usepackage{amsthm}%
\usepackage{mathrsfs}%
\usepackage[title]{appendix}%
\usepackage{xcolor}%
\usepackage{textcomp}%
\usepackage{manyfoot}%
\usepackage{booktabs}%
\usepackage{algorithm}%
\usepackage{amssymb}
\usepackage{algorithmicx}%
\usepackage{algpseudocode}%
\usepackage{listings}%
\usepackage[utf8]{inputenc}
\usepackage{amsmath}
\usepackage[utf8]{inputenc}
\usepackage{newunicodechar}

\theoremstyle{thmstyleone}%
\theoremstyle{thmstyletwo}%

\theoremstyle{thmstylethree}%

\begin{document}
\title[Article Title]{VRAE: Vertical Residual Autoencoder for License Plate Denoising and Deblurring}


\author[1]{\fnm{Cuong Nguyen}}\email{cuong.nt227090@sis.hust.edu.vn}

\author[2]{\fnm{Dung T. Tran}}\email{dung.tt2@vinuni.edu.vn}

\author[3]{\fnm{Hong Nguyen}}\email{hongn@usc.edu}
\author[1]{\fnm{Xuan-Vu Phan}}\email{vu.phanxuan@hust.edu.vn}
\author*[1]{\fnm{Nam-Phong Nguyen}}\email{phong.nguyennam@hust.edu.vn}

\affil[1]{\orgname{Ha Noi University of Science and Technology}, \orgaddress{\city{Ha Noi}, \country{Vietnam}}}

\affil[2]{\orgdiv{Center of Environmental Intelligence}, \orgname{VinUniversity}, \city{Hanoi}, \country{Vietnam}}

\affil[3]{\orgname{University of Southern California}, \orgaddress{\city{Los Angeles}, \country{United State}}}


\abstract{In real-world traffic surveillance, vehicle images captured under adverse weather, poor lighting, or high-speed motion often suffer from severe noise and blur. Such degradations significantly reduce the accuracy of license plate recognition systems, especially when the plate occupies only a small region within the full vehicle image. Restoring these degraded images a fast realtime manner is thus a crucial pre-processing step to enhance recognition performance. In this work, we propose a Vertical Residual Autoencoder (VRAE) architecture designed for the image enhancement task in traffic surveillance. The method incorporates an enhancement strategy that employs an auxiliary block, which injects input-aware features at each encoding stage to guide the representation learning process, enabling better general information preservation throughout the network compared to conventional autoencoders. Experiments on a vehicle image dataset with visible license plates demonstrate that our method consistently outperforms Autoencoder (AE), Generative Adversarial Network (GAN), and Flow-Based (FB) approaches. Compared with AE at the same depth, it improves PSNR by about 20\%, reduces NMSE by around 50\%, and enhances SSIM by 1\%, while requiring only a marginal increase of roughly 1\% in parameters.}



\maketitle

\section{Introdution \& Related Work}
\label{sec:Introdution}
In recent years, traffic scene understanding has become essential for intelligent transportation, autonomous driving, and surveillance. However, real-world traffic images are often degraded, reducing system reliability. Motion blur frequently occurs due to fast vehicle–camera motion and low-cost sensors with limited exposure, while advanced high-speed cameras remain costly~\cite{pan2016dark,kupyn2018deblurgan,nah2017deep}. Moreover, low-light conditions further introduce noise, glare, and low contrast, particularly in Southeast Asian cities with dense nighttime traffic and mixed lighting \cite{chen2018see,jiang2019enlightengan,zamir2020learning}. In addition, sensor limitations such as noise and dynamic range exacerbate the trade-off between blur and signal strength \cite{zhang2017beyond,anwar2020real}. As a result, these degradations severely impair downstream tasks like detection, recognition, and segmentation \cite{redmon2018yolov3,he2017mask}. To address this, image enhancement techniques, especially including denoiser and deblurrer, are widely studied. While traditionally treated separately, traffic images often suffer from both simultaneously. Deep learning methods have been explored including AE for compact representations, GAN for enhanced perceptual quality, and FB for explicit distribution learning that balances fidelity and diversity.

Traditional methods using AE have been widely adopted in such tasks due to their ability to learn compact latent representations and reconstruct images by filtering noise from signal \cite{vincent2008extracting}. Classical convolutional AE have shown promising results in removing additive noise, as demonstrated in DnCNN \cite{zhang2017beyond}, which employed residual learning and batch normalization for Gaussian denoising. Similarly, convolutional encoder-decoder structures have been successfully applied in deblurring, such as DeblurGAN \cite{kupyn2018deblurgan}, which combined adversarial training with perceptual loss to recover sharp textures. This facilitates direct mapping from corrupted inputs to clean outputs, effectively addressing complex or spatially varying degradations.
State-of-the-art deep generative methods, such as GAN and Diffusion, have been integrated into restoration pipelines. Specifically, DeblurGAN-v2 \cite{kupyn2019deblurganv2} introduced a multi-scale GAN framework that achieved both sharpness and stability in deblurring.
Recently, FB models have emerged as a principled alternative for image restoration, capable of learning invertible mappings between degraded and clean domains while preserving density information \cite{kingma2018glow}.Notably, FB has been introduced as a scalable and stable training framework for conditional image generation, with its application to restoration explored in tasks demanding spatial consistency and smooth transitions, such as PnP-Flow \cite{martin2025pnpflow,zhu2024flowie}.

\textbf{Research Gap:} AE have been employed in traffic image enhancement due to their lightweight design, which is suitable for resource-constrained monitoring systems~\cite{vincent2008denoising,xu2020lightweight,jiang2018deep}. However, shallow encoders typically capture only low-level features and fail to recover mid- and high-level structures that are critical for fine detail restoration \cite{bengio2013representation,zhang2018srmd}. GAN improve perceptual realism with sharper edges and more natural textures, but often distort pixel-level fidelity, hallucinate numbers, limiting their reliability for traffic applications where accurate recovery of details such as license plates and road markings is essential \cite{ledig2017photo,wang2018esrgan}. FB models preserve pixel-level accuracy by optimizing explicit likelihood functions, yet their high computational complexity and memory consumption make them impractical for real-time deployment on edge devices \cite{kingma2018glow,zhang2021invertible}. Therefore, a clear research gap exists in developing architectures that balance computational efficiency with accurate detail preservation for traffic image enhancement.
For such reasons, the main contributions of this work can be summarized as follows:
\begin{itemize}
    \item We propose the \textbf{Vertical Residual Autoencoder (VRAE)} with an auxiliary block that vertically aggregates and refines embeddings for enhanced feature representation. Experimental results show that VRAE achieves impressive improvements across PSNR, SSIM, and NMSE, while maintaining competitive performance even with reduced model parameters.
    \item Through extensive experiments, we show that our approach not only encourages general information preservation in the early encoder layers but also outperforms conventional autoencoders in terms of image compression, as evaluated by entropy change
\end{itemize}

\begin{figure}[htbp]
    \centering
    \includegraphics[width=1\textwidth]{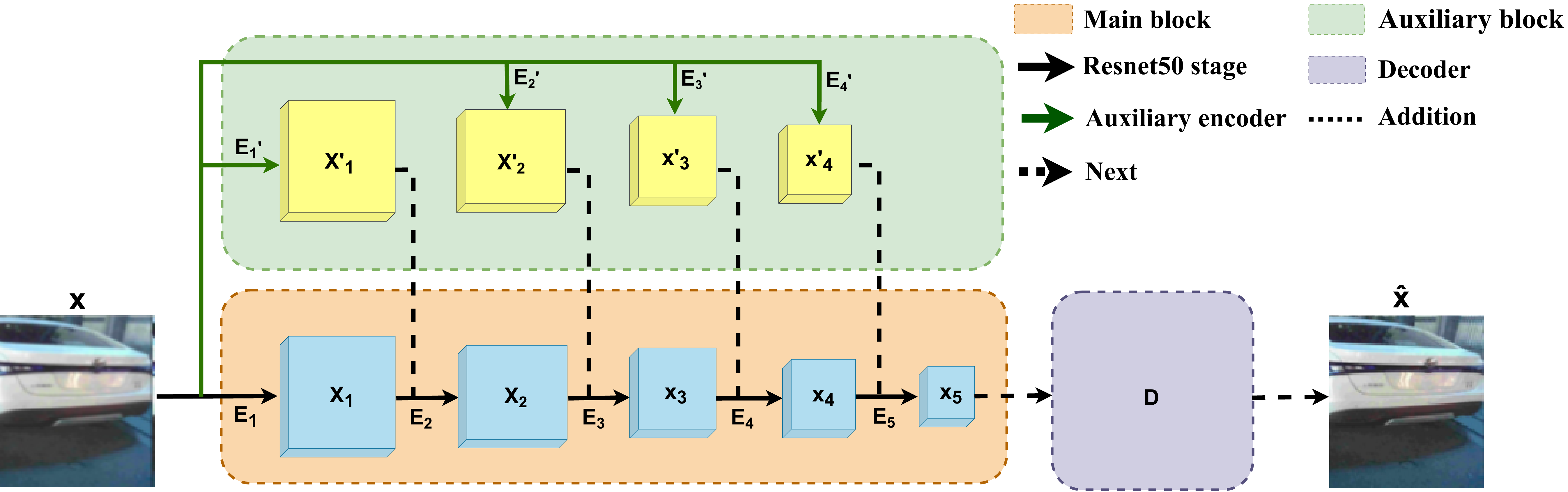}
    \caption{VRAE aggregates the input embedding by passing it through the feature embedding block before forwarding it to the next encoder block.}
    \label{fig:VRAE}
\end{figure}
\section{Vertical Residual Autoencoder }
\textbf{Problem statement}
In traffic surveillance systems, vehicle images are often degraded by motion blur, sensor noise or adverse weather conditions, which obscure important visual details such as license plates, headlights, and road markings. Assume that all distortion is created by $\epsilon \sim N(0, 1)$ and the task of vehicle image enhancement is defined as learning a one-to-one mapping $\mathbb{F}\colon \mathbf{x} \rightarrow \hat{\mathbf{x}}$ from a degraded input image $\mathbf{x} \in \mathbb{R}^{3 \times W \times H}$ to an enhanced output image $\hat{\mathbf{x}} \in \mathbb{R}^{3 \times W \times H}$, where $\mathbf{x} = \hat{\mathbf{x}} + \epsilon$.


\subsection{Overall Architecture}\label{sec:overall}
VRAE comprises three components: (i) a parallel stack of \emph{Auxiliary Encoders} $\{E_i\}_{i=1}^{N}$, each extracting hierarchical features directly from the input $x$; (ii) a continuous stack of \emph{Main Encoders} $\{E'_i\}_{i=1}^{N-1}$ that further transforms the intermediate features and embeds them for residual injection into deeper stages; and (iii) a \emph{Decoder} $D$ that aggregates features from both streams to reconstruct the output $\hat{x}$. The two streams interact via element-wise additions at multiple stages, enabling vertical residual learning.
\subsection{Auxiliary Block}\label{sec:aux-encoder}
The auxiliary path $\{E'_i\}$ acts as a \emph{feature embedding module} that repeatedly extracts complementary, input-conditioned features to be fused with the main stream. Each block follows a shallow Conv--Norm--Activation structure and optionally includes pooling to summarize global context:
\begin{equation}
    \mathbf{x}'_{i} = E'_i(\mathbf{x}), \quad i=1,\dots,5.
\end{equation}
As illustrated in Fig.~\ref{fig:VRAE}, the auxiliary block operates in parallel with the main encoder and generates an additional feature hierarchy $\{\mathbf{x}'_{i}\}$ from the degraded input $\mathbf{x}$. These features are progressively injected into the corresponding layers of the main block through element-wise fusion, allowing the model to enrich low- and mid-level representations with complementary cues. In particular, the auxiliary branch facilitates disentangling noise and blur from meaningful structures, thereby enhancing the main encoder’s capacity to capture fine details that are otherwise lost in shallow representations.
\subsection{Main Block}\label{sec:main-encoder}
In our implementation, each main encoder $E_i$ is instantiated as the $i$-th stage of the ResNet-50 architecture~\cite{he2016deep}.  
ResNet-50 employs a bottleneck design consisting of three convolutional layers: a $1\times1$ convolution for dimensionality reduction, a $3\times3$ convolution for spatial processing, and another $1\times1$ convolution for restoring dimensions. A skip connection adds the input to the output of the block, enabling better gradient flow and alleviating the vanishing gradient problem.  

At each stage, the feature maps from the main and auxiliary encoder paths are fused by element-wise addition before being passed into $E_i$. Formally:
\begin{equation}
    \mathbf{x}_{i} = E_{i}\!\left(\mathbf{x}_{i-1} + \mathbf{x}'_{i-1}\right), 
    \quad i = 2,\dots,5,
\end{equation}
where $\mathbf{x}_{i-1}$ is the output of the $(i-1)$-th main encoder block and $\mathbf{x}'_{i-1}$ is the output of the $(i-1)$-th auxiliary encoder block.  
This additive fusion serves as an embedding-level integration mechanism: it incorporates complementary information without increasing feature dimensionality, offering a parameter-efficient alternative to concatenation. Such design, inspired by residual networks~\cite{he2016deep} and multi-branch models~\cite{huang2017densely}, not only stabilizes gradient flow but also enhances representational power, allowing the main encoder to capture hierarchical abstractions while continuously integrating localized cues from the auxiliary encoder.

\subsection{Decoder}\label{sec:decoder}
The decoder $D$ reconstructs $\hat{\mathbf{x}}$ from $\mathbf{x}_N$ using a cascade of \emph{transposed convolutional layers} followed by \emph{ReLU} activations. Each transposed convolution progressively upsamples the feature maps, restoring the spatial resolution:
\begin{equation}
    \mathbf{y}_{k} = \mathrm{ReLU}\!\left(\mathrm{ConvTranspose2d}_k(\mathbf{y}_{k-1})\right), 
    \quad k=1,\dots,5,\ \ \mathbf{y}_0=\mathbf{x}_N,\ \ \mathbf{y}_K=\hat{\mathbf{x}}.
\end{equation}

\noindent
We train the network with the \emph{mean squared error (MSE)} loss only:
\begin{equation}
\mathcal{L}_{\mathrm{MSE}}
= \frac{1}{BCHW}\!\sum_{b=1}^{B}\sum_{c=1}^{C}\sum_{i=1}^{H}\sum_{j=1}^{W}
\big(\hat{x}_{b,c,i,j}-x_{b,c,i,j}\big)^{2},
\end{equation}
which averages the squared reconstruction error over the entire batch and all pixels/channels. 

\section{Experiments}

\subsection{Dataset Preparation}
From the CCPD dataset \cite{xu2018towards}, 3,036 high-resolution vehicle images were collected and resized to $256 \times 256$ pixels. The dataset was split into training, validation, and test sets in a 70\%–15\%–15\% ratio. Data augmentation, including rotations, was applied only to the training set, increasing it to 7,000 images. Low-resolution inputs were generated by adding discrete noise (values in $[0, 9]$ scaled by 0.1) and applying a $3 \times 3$ average pooling filter iteratively 10 times, simulating low-quality surveillance footage. This degradation process is not identical but bears similarity to the synthetic degradations used in SRMD \cite{zhang2018srmd}, where combinations of blurring and noise are employed to simulate multiple low-resolution conditions.

\subsection{Baseline choices}
In vehicle image restoration, GAN have been widely applied to inpainting, deblurring, super-resolution, and artifact removal, achieving visually plausible and high-quality results \cite{ledig2017photo,isola2017pix2pix,kupyn2019deblurganv2}. 
In contrast, flow-based models such as RealNVP \cite{dinh2017realnvp} and Glow \cite{kingma2018glow} provide exact invertibility between image and latent spaces, enabling better preservation of fine-grained details. 
Although still underexplored in this domain, these characteristics make flow-based approaches particularly relevant to vehicle image restoration. 
To ensure a fair and controlled comparison, in this work we select both GAN and FB methods as baselines and design them with main encoder and decoder blocks consistent with our proposed method.
\subsection{Settings}
All models were trained using the Adam optimizer with an initial learning rate of $1\times10^{-4}$ and a batch size of 16. 
Each model was trained for 100 epochs on an NVIDIA RTX 4070 GPU.

\begin{itemize}
    \item \textbf{AE, FB, VRAE}: These methods utilize the MSE loss between the predicted and ground-truth HR images

    \item \textbf{GAN}: The GAN generator is trained with a hybrid loss combining pixel-wise MSE and adversarial BCE. While MSE ensures closeness to the ground truth, it often leads to overly smooth outputs. The adversarial loss encourages perceptual realism by pushing the generator to fool the discriminator. The overall generator loss is thus:
\[
\mathcal{L}_{\text{GEN}} = \mathcal{L}_{\text{MSE}} + \alpha \cdot \mathcal{L}_{\text{BCE}}
\]
where $\alpha = 10^{-3}$ is a small scalar balancing the adversarial loss contribution, following the original setting proposed in \cite{ledig2017photo}.

\end{itemize}


\begin{table}[t]
\centering
\normalsize
\caption{Quantitative comparison of GAN, FB, AE (2--5 sub encoders), 
and the proposed VRAE (2--5 sub-encoders). 
All metrics are evaluated on the \textbf{test set} and averaged across all samples. 
The best results are in bold, and the second-best are underlined.}
\label{tab:results}
\begin{tabular}{lccccc}
\toprule
\textbf{Model} & \textbf{PSNR ($\uparrow$)} & \textbf{NMSE ($\downarrow$)} & \textbf{SSIM ($\uparrow$)} & \textbf{Params} & \textbf{FPS} \\
\midrule
GAN (Gen)      & 28.743  & 0.006 & 0.842 & 14.59M    & 188 \\
FB  (Gen)      & 27.093  & 0.008 & 0.831 & 25.87M    & 35  \\
AE2            & 27.787  & 0.007 & 0.840 & \textbf{0.375M}\footnotemark[1] & \textbf{411} \\
AE3            & 26.159  & 0.011 & 0.802 & 2.77M     & 289 \\
AE4            & 29.404  & 0.005 & 0.864 & 14.59M    & 189 \\
AE5            & 27.243  & 0.008 & 0.793 & 48.43M    & 119 \\
VRAE2          & \underline{30.319} & \underline{0.004} & \underline{0.884} & \underline{0.376M}   & \underline{399} \\
VRAE3          & \textbf{31.052}    & \textbf{0.003}    & \textbf{0.898}    & 2.78M     & 194 \\
VRAE4          & 30.246  & 0.004 & 0.880 & 14.61M    & 90  \\
VRAE5          & 30.032  & 0.003 & 0.860 & 48.48M    & 45  \\
\bottomrule
\end{tabular}
\footnotetext[1]{M denotes millions, the unit of the number of parameters.}
\end{table}
\section{Result}
\subsection{Comparison with State-of-the-Art}
Table~\ref{tab:results} presents the quantitative comparison among GAN, FB, AE (2--5 stacks), and the proposed VRAE (2--5 sub-encoders). 
Conventional GAN-based and FB-based approaches show moderate perceptual quality but achieve relatively lower PSNR and SSIM values compared to AE-based and VRAE-based models. 
Although stacked Autoencoders (e.g., AE4) can provide competitive results (PSNR = 29.404 dB, SSIM = 0.864), their performance is inconsistent across different stack depths, and some configurations (e.g., AE3 and AE5) degrade significantly. 
In contrast, the proposed VRAE consistently outperforms both AE and GAN variants in terms of reconstruction quality. 
Notably, VRAE3 achieves the highest PSNR (31.052 dB) and SSIM (0.898), while also maintaining a low NMSE (0.003). 
VRAE2 also delivers strong results (PSNR = 30.319 dB, SSIM = 0.884), ranking second overall. 
Despite slightly higher parameter counts than AE2, VRAE models demonstrate a favorable trade-off between accuracy and efficiency, with FPS values remaining competitive (e.g., VRAE2 at 399 FPS versus AE2 at 411 FPS). 
When compared with AE at the same stack depth, VRAE improves PSNR by approximately 20\%, reduces NMSE by around 50\%, and enhances SSIM by about 1\%, while introducing only a marginal increase of roughly 1\% in the number of parameters. 
These results indicate that the VRAE architecture not only stabilizes performance across different depths but also surpasses conventional baselines in both reconstruction fidelity and computational efficiency.

\subsection{Informative content preservation}
\begin{figure}[tp]
    \centering
    \includegraphics[width=1\textwidth]{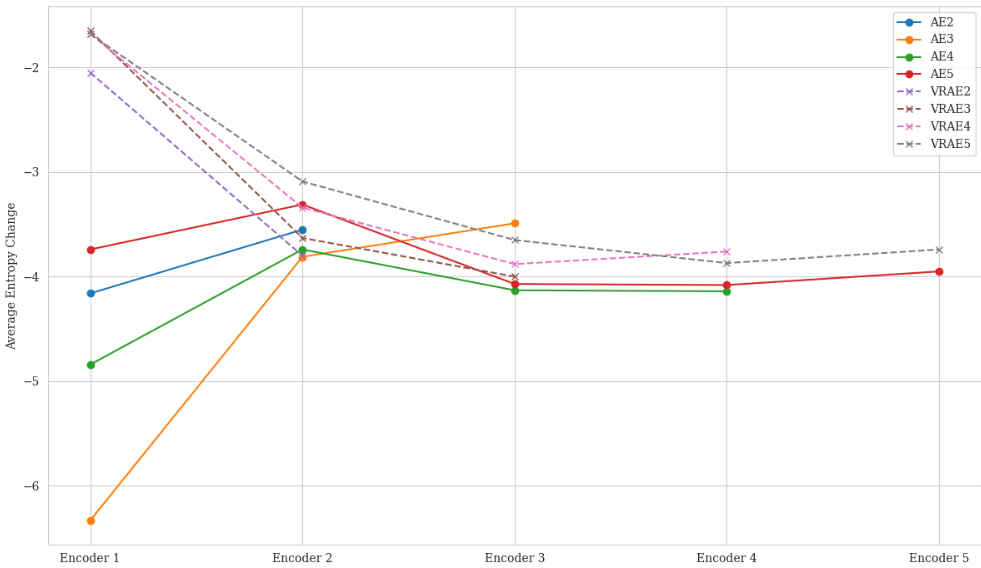}
    \caption{Average entropy change across encoder layers of AE and VRAE.}
    \label{fig:entropy_dynamics}
\end{figure}
According to \cite{entropy_guidance}, encouraging entropy preservation in the early layers promotes better generalization. This is consistent with our hypothesis Figure~\ref{fig:entropy_dynamics}, where we observe that models preserving higher entropy in the first encoder layers tend to achieve better stability across subsequent layers.
We first define the feature maps at encoder layer $l$ as
\begin{equation}
    F_l \in \mathbb{R}^{C_l \times H_l \times W_l},
\end{equation}
where $C_l$ is the number of channels, and $H_l, W_l$ are the spatial height and width of the feature maps.
Formally, the entropy change between consecutive layers $l$ and $l+1$ is defined as
\begin{equation}
    \Delta H_{l} = H(F_{l+1}) - H(F_{l}),
\end{equation}
where $H(F_l)$ denotes the Shannon entropy of the feature maps $F_l$. The Shannon entropy is defined as
\begin{equation}
    H(X) = - \sum_{i} p(x_i) \, \log p(x_i),
\end{equation}
where $X$ is a random variable, $x_i$ are possible outcomes (here corresponding to activation values in feature maps), and $p(x_i)$ is the empirical probability of observing $x_i$.
According to the proof in \cite{entropy_guidance}, this entropy change can be approximated by the following proxy formulation:
\begin{equation}
    \Delta H_{l} = (h - p + 1)(w - q + 1) \cdot \log \big|c_{11}\big|,
\end{equation}
where $h$ and $w$ denote the height and width of the input feature map, 
$p$ and $q$ represent the height and width of the convolutional kernel, 
and $c_{11}$ refers to the top-left coefficient of the convolution filter.
This proxy captures how convolutional kernels affect the representational diversity of feature maps. A positive $\Delta H_l$ implies increased representational diversity, while a negative $\Delta H_l$ indicates information compression. In this work, we extend the formulation to operate at the encoder--block level instead of individual convolutional layers. Given an encoder block $B_k$ containing $m$ convolutional layers, we compute the average entropy change as:
\begin{equation}
    \Delta H^{(B_k)}_{\text{avg}} = \frac{1}{m} \sum_{l \in B_k} \big[ H(F_{l+1}) - H(F_{l}) \big].
\end{equation}
This block-wise aggregation smooths local fluctuations and enables a more stable comparison across architectures and depths.
We now provide a deeper analysis of the entropy change patterns observed in Figure~\ref{fig:entropy_dynamics}.

\noindent
The first layer exhibits the largest discrepancy in entropy change between AE and VRAE models. For instance, AE3 experiences a sharp entropy reduction at Encoder 1, indicating a substantial loss of information at the very beginning. In contrast, VRAE models start with higher entropy change, suggesting that they preserve richer information in the initial stage. This aligns with the hypothesis that maintaining entropy early helps avoid premature information collapse.

\noindent
As we move through the encoder layers, AE models often exhibit stronger fluctuations in entropy reduction. Some autoencoders show sharp drops at specific layers, while others stabilize more gradually. On the other hand, VRAE models generally demonstrate a smoother and more consistent entropy decrease. This gradual decline implies that VRAEs manage information compression more effectively, reducing the risk of losing essential features too quickly, as evidenced by their smoother entropy change across layers—for instance, VRAE3 decreases from approximately –1.5 at the first encoder to –3.7 at the second (a change of about 2.2 unit), whereas AE3 drops from around –6.3 to –3.5 in the same transition (a change of about 2.8 unit).

\noindent
In the later encoder layers (Encoder 4–5), both AE and VRAE models tend to converge toward similar entropy levels. However, AE variants usually stabilize at lower entropy values, meaning they have compressed the information more aggressively. VRAE models, by contrast, maintain slightly higher entropy in the deeper layers, reflecting their ability to retain more useful representational capacity.
Overall, these results support the claim that maintaining entropy diversity in earlier layers helps the network preserve useful representational capacity.\\

\begin{figure}[tp]
    \centering
    \includegraphics[width=1\textwidth]{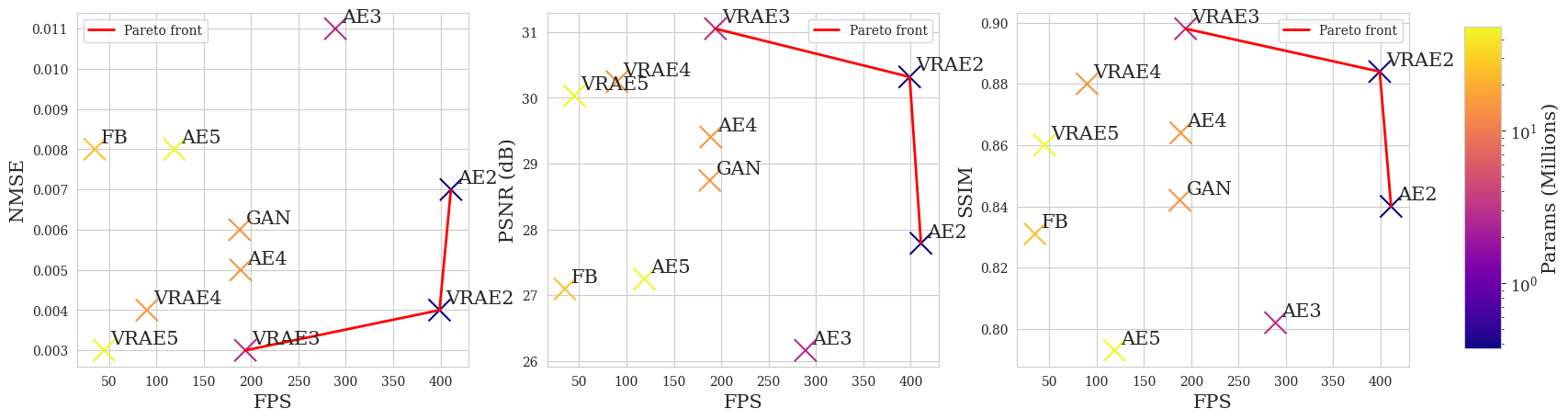}
    \caption{Visualization of model performance trade-offs using the Pareto front. 
Each point denotes a model evaluated by reconstruction quality (NMSE, PSNR, SSIM) versus inference speed (FPS), 
with color indicating model size, i.e., the number of parameters. 
The red line represents the Pareto front, i.e., the set of models that cannot be improved in one objective without sacrificing another. 
This highlights the optimal balance between accuracy and efficiency, helping to identify the most suitable models depending on whether speed, quality, or a balanced trade-off is prioritized.}
    \label{fig:paretoline}
\end{figure}
\subsection{Comparative analysis of model trade-offs}
To evaluate the efficiency of different models in vehicle image restoration, we compare their performance in terms of quality metrics (PSNR, SSIM, and NMSE) against inference speed (FPS). Figure~\ref{fig:paretoline} shows the distribution of models with the Pareto front highlighted in red, while the color scale represents the number of parameters in millions.

In the PSNR--FPS and SSIM--FPS plots, we observe that VRAE-based models (e.g., VRAE3 and VRAE2) consistently lie on the Pareto front, striking a balance between image quality and computational speed. In contrast, shallow autoencoders (e.g., AE2 and AE3) achieve high FPS but suffer from significantly lower quality scores, indicating that prioritizing speed alone comes at the cost of detail preservation. Meanwhile, GAN and flow-based (FB) models demonstrate moderate performance, but their relatively high parameter counts make them less efficient.

The NMSE--FPS trade-off further highlights these limitations: although AE2 reaches the highest FPS, it exhibits considerably larger reconstruction error, while deeper VRAE variants (VRAE3, VRAE2) maintain low NMSE values with acceptable inference speeds. These results underline the importance of designing models that preserve sufficient depth of encoding to capture meaningful representations while avoiding excessive parameter growth.

Overall, the Pareto front demonstrates that VRAE architectures achieve a favorable trade-off across accuracy, speed, and parameter efficiency, making them strong candidates for real-time vehicle image restoration on edge devices.

\subsection{Limitations}
Although the proposed VRAE models achieve superior reconstruction quality in terms of PSNR, NMSE, and SSIM compared to conventional baselines, they also reveal a notable limitation. As shown in Table~\ref{tab:results}, deeper VRAE variants (e.g., VRAE4 and VRAE5) provide consistent quality improvements but at the expense of significantly increased parameter counts and reduced inference speed. This highlights a trade-off: obtaining higher-quality restoration requires substantially more training resources and memory, which may limit their deployment on resource-constrained edge devices. Therefore, identifying an optimal balance between accuracy and efficiency remains a critical challenge for future work.
\section{Conclusion and future work}
VRAE was proposed the Variational Residual Autoencoder (VRAE) for vehicle image restoration, which combines an auxiliary encoder and feature aggregation to enhance local detail and hierarchical representation. VRAE outperforms conventional Autoencoders, GANs, and flow-based models in PSNR, SSIM, and NMSE, while offering competitive inference speed. However, deeper VRAE variants improve quality at the cost of higher complexity, limiting edge deployment. Future work includes exploring lightweight designs, compression methods, e.g., pruning, quantization, distillation, and adaptive-depth architectures. VRAE can be further extended to real-time video restoration and other intelligent transportation tasks in future work.
\bibliography{references}
\end{document}